\documentclass[10pt, a4paper]{article}

\usepackage[final]{lrec2026} 

\title{Multimodal Analysis of State-Funded News Coverage of the Israel–Hamas War on YouTube Shorts}

\name{{\textbf{Daniel Miehling}} and {\textbf{Sandra Kübler}}}

\address{Indiana University\\
         \{damieh, skuebler\}@iu.edu\\}

\abstract{YouTube Shorts have become central to news consumption on the platform, yet research on how geopolitical events are represented in this format remains limited. To address this gap, we present a multimodal pipeline that combines automatic transcription, aspect-based sentiment analysis (ABSA), and semantic scene classification. The pipeline is first assessed for feasibility and then applied to analyze short-form coverage of the Israel–Hamas war by state-funded outlets. Using over 2,300 conflict-related Shorts and more than 94,000 visual frames, we systematically examine war reporting across major international broadcasters. Our findings reveal that the sentiment expressed in transcripts regarding specific aspects differs across outlets and over time, whereas scene-type classifications reflect visual cues consistent with real-world events. Notably, smaller domain-adapted models outperform large transformers and even LLMs for sentiment analysis, underscoring the value of resource-efficient approaches for humanities research. The pipeline serves as a template for other short-form platforms, such as TikTok and Instagram, and demonstrates how multimodal methods, combined with qualitative interpretation, can characterize sentiment patterns and visual cues in algorithmically driven video environments.
 \\ \newline  }

\begin{document}

\maketitleabstract

\section{Introduction}
Short-form video has become a dominant format for political communication on platforms such as YouTube, TikTok, and Instagram \cite{guinaudeau_et_al}, and it plays an increasingly important role in conflict coverage. Short videos compress complex events into simple narratives that prioritize audience reaction over contextual nuance, which may distort the representation while still being highly persuasive. They often appear in users’ timelines with little information about the creator’s ideological stance or partisan alignment. In recommendation-driven environments, multimodal signals such as text, visuals, and audio can amplify emotional framing and potentially shape audience perception \cite{Karduni_Wesslen_Markant_Dou_2023, arora-etal-2025-multi, Narkar_Vohra_KhudaBukhsh_2025}, although some research disputes this claim \cite{liu_et_al}. 

Following Hamas’s attacks on Israel on October~7, 2023, online news coverage was accompanied by an explosion of user-generated content \citep{alamsyah2024, miner-ortega-2024-nlp, steffen2025memesmultimodaltopicmodeling}.
Despite this trend, little is known about how (geo-)political events are represented in short-form news, particularly by state-funded outlets. Prior work shows that multimodal signals shape online environments: \citet{mynt} jointly model sentiment, emotion, and attention words, while \citet{KUMARI2021102631} combine misinformation detection, emotion recogntion and novelty detection.
Research on TikTok highlights similar dynamics: \citet{kriss2024fun} show that emotionally blended messages increase engagement, and \citet{chenzi} demonstrate that negative sentiment and second-person address elicit stronger audience responses.

Our work builds on this prior research and investigates how state-funded outlets cover the Israel--Hamas war and how they use sentiment and visual scene types to flesh out their narrative.

\section{Related Work}
We build on prior work in multimodal analysis of social media by applying these approaches directly to short-form video content, complementing research that examines user responses to such material. Several multimodal frameworks provide methodological foundations for integrating textual, visual, and audio cues. The MultiTec framework \cite{shang2025multitec} fuses ASR, OCR, visual features, audio sentiment, and metadata to investigate Healthcare misinformation on TikTok. \citet{ali2023multimodal} use multimodal critical discourse analysis to show how TikTok videos combine visual elements, on-screen text, and gestures to construct ideological narratives. \citet{Gong_Mousavi_Xia_Zannettou_2025} introduce ClipMind, a multimodal framework for evaluating AI-based algorithmic recommendation systems on TikTok. The approach generates bimodal representations of visual and audio content to differentiate mainstream topics such as food and beauty care from niche interests including war and mental health.  \citet{Rizwan_Bhaskar_Das_Majhi_Saha_Mukherjee_2025} explore VLMs to detect hateful content in memes, showing that joint reasoning over text and images outperforms unimodal baselines for capturing implicit and explicit hate speech. 
\citet{chatterje-doody2019making} show that multimodal representations shape audience interpretation in conflict reporting by combining narrative and visual structures with affective cues.

Prior research on the Israel–Hamas conflict has largely focused on user-generated content \cite{becker2023celebrating, jikeli2023holocaust}, with subsequent work highlighting differences between pro-Israel and pro-Palestine discourse using network analysis and aspect-based sentiment analysis of user comments \cite{alamsyah2024, miehling-dakota-kbler:2025:RANLP}. \citet{zaghouani-etal-2024-fignews} introduce multilingual bias and propaganda annotation for early Israel–Hamas war news. Related work has also examined the interaction of visual cues and text in online memes, demonstrating how bimodal relationships structure meaning \cite{steffen2025memesmultimodaltopicmodeling}. In contrast, we analyze short-form news videos themselves, combining transcript-based ABSA with visual scene-type classification to examine how state-funded broadcasters portray (geo-)political actors in short-form conflict coverage.

\section{Research Question}

Building on work by \citet{miehling-dakota-kbler:2025:RANLP}, we introduce a multimodal pipeline that combines automatic transcription, dependency-based aspect linking, aspect-based sentiment analysis (ABSA), and visual scene-type classification, along with a taxonomy for war-related visual frames. We apply this framework to more than 2,300 YouTube Shorts collected over one year to analyze how state-funded outlets cover the Israel–Hamas war between 2023 and 2024, comparing sentiment and visual scene distributions across outlets and over time. This study addresses the following research question:

\textbf{RQ:} How do state-funded or state-supported outlets differ in their use of sentiment and visual scene types when covering key (geo-)political actors?

\section{Semantic Scene Analysis}
To assess visual content in Shorts, we have developed a novel taxonomy of scene types. While prompt-based VLM can generate rich open-ended image descriptions, these descriptions are difficult to aggregate systematically at scale. Our taxonomy categorizes scenes into a small set of types, which were developed using an iterative refining process iterating over portions of our dataset. The final taxonomy comprises seven distinguishable semantic scene types with refined definitions that minimize overlap while covering dominant visual themes in 2023--2024 Israel--Hamas war coverage.

We define the visual scene types shown in Appendix~\ref{appendix:prompt} based on observable cues commonly used in war reporting, using conservative assignment rules to minimize ambiguity.

\begin{figure}[t]
\centering
\includegraphics[width=0.48\textwidth]{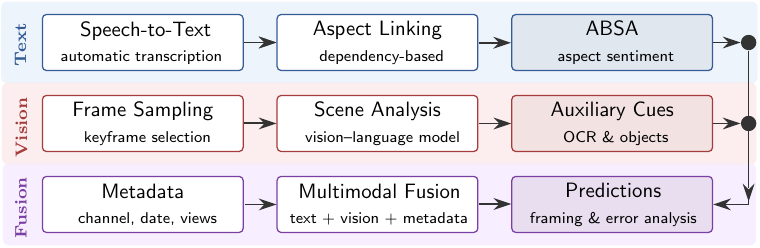} 
\caption{Multimodal Pipeline for Shorts Analysis.}
\label{fig:p}
\end{figure}

\section{Methodology}
We propose a multimodal framework that integrates \texttt{Whisper}-generated transcripts \cite{whisperopenai}, computer vision for semantic scene classification \cite{qwen3vl_2025}, and Aspect-Based Sentiment Analysis (ABSA) \cite{YangPyABSA, he2021debertav3} to examine how specific aspect categories are referenced and evaluated in short-form video content. 
Figure~\ref{fig:p} illustrates the pipeline\footnote{See the project repository for the full code: \url{https://github.com/damieh1/YTSC}}.

\subsection{Data Collection}
The dataset was constructed using the official YouTube API\footnote{\url{https://developers.google.com/youtube/v3/docs}}, beginning two weeks after October~7, 2023. The analysis focuses on state-funded or state-supported international broadcasters. Over twelve months, 2{,}371 YouTube Shorts were collected from four channels, Al Jazeera, BBC, Deutsche Welle (DW), and TRT World. Although all outlets primarily report in English, 36 Shorts contained predominantly non-English content and were excluded from the linguistic analysis, resulting in a final set of 2{,}335 Shorts. For details, see Table~\ref{tab:spoken-stats}. The shorts were manually selected and scraped approximately two weeks after publication. 

\subsection{Speech-to-Text}
We used \texttt{Whisper large-v3} \cite{whisperopenai} to generate textual transcripts. We first checked each video for up to 30s to infer or validate the dominant spoken language before full transcription.
A small manual evaluation showed superior punctuation, segmentation, and robustness compared to YouTube auto-captions.

\begin{table}[t]
\centering
\setlength{\tabcolsep}{4pt}
\begin{tabular}{lrrrr}
Source      & Video IDs & Spoken & Sp.\ \% & No sp. \\
\hline
Al~Jaz.\  & 924   & 838  & 90.7   & 86 \\
BBC         & 68    & 68   & 100   & 0              \\
DW          & 85    & 83   & 97.6  & 2              \\
TRT~World         & 1{,}258  & 1{,}249 & 99.2   & 9              \\
\hline
\end{tabular}
\caption{Descriptive statistics for the transcript corpus. No speech (No Sp.) indicates visual-only or non-verbal audio content.}
\label{tab:spoken-stats}
\end{table}

In total, we processed all videos with mainly English content of the four outlets. From Al~Jazeera (AJ), 924 Shorts were processed; 838 (90.7\%) produced usable transcripts, whereas 86 (9.3\%) consisted solely of visual coverage or background sound without intelligible speech. For TRT, 1{,}249 Shorts were reviewed; 99.2\% contained spoken language, and 9 were visual-only. Detailed counts are provided in Table~\ref{tab:spoken-stats}.

\begin{table}[t]
\centering
\begin{tabular}{lrr}
Outlet & Examples & Unique Videos \\
\hline
Al Jazeera & 1{,}614 & 393 \\
BBC & 236  & 41  \\
DW  & 128  & 28  \\
TRT~World & 1{,}274 & 320 \\
\hline
Total & 3{,}252 & 782 \\
\hline
\end{tabular}
\caption{Dependency-anchored aspects per outlet.}
\label{tab:dep-parse-counts}
\end{table} 

\begin{figure}[t]
\centering
\includegraphics[width=1.0\columnwidth]{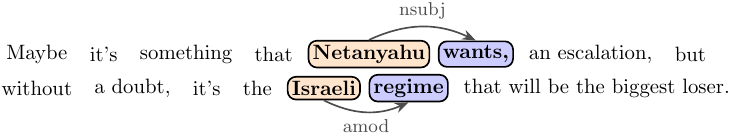}
\caption{Dependency analysis for aspects 'Netanyahu' and 'Israeli'. Predicted Sentiment: Neg. (0.99).} \label{parser}
\end{figure}

\begin{table}[t]
\centering
\begin{tabular}{lrrrr}
Aspect Group & Neg & Neut & Pos & Total \\
\hline
Arab & 0 & 26 & 2 & 28 \\
Gaza & 141 & 243 & 183 & 567 \\
Islam & 112 & 279 & 234 & 625 \\
Islamism & 612 & 570 & 244 & 1426 \\
Israel & 551 & 671 & 398 & 1620 \\
Israeli politicians & 34 & 38 & 1 & 73 \\
Jews & 199 & 272 & 110 & 581 \\
Opp.\ non-Isr.\ polit.\ & 0 & 11 & 2 & 13 \\
Palestine & 452 & 623 & 442 & 1517 \\
Zion & 232 & 274 & 25 & 531 \\
\hline
\end{tabular}
\caption{Sentiment label distribution in the gold-standard training data by canonical aspect group.} 
\label{tab:canonical-absa}
\end{table}

\subsection{Aspect Linking}
We converted the \texttt{Whisper}-generated transcripts into structured, syntax-anchored aspect rows using the dependency parser by \citet{dozat:manning:17} (model: \texttt{biaffine-dep-en}), implemented in the \texttt{SuPar} library. The parser was trained on English Universal Dependencies (UD) treebanks. From the parses, we extracted dependency triples (for more details, see below).

Our aspect lexicon consists of 56 manually selected surface forms grouped into ten substantive categories relevant to the Israel--Hamas war: \textit{Israel}, \textit{Palestine}, \textit{Hamas/Islamism}, \textit{Jews}, \textit{Muslims/Islam}, and \textit{Zionism}, along with named political actors (e.g., \textit{Netanyahu}, \textit{Sinwar}). Each category contains multiple lexical variants (see Appendix~\ref{appendix:lexicon} for the complete lexicon). 

Using this lexicon, we iterated over all transcript segments and retained only those sentences containing at least one aspect match. We parsed each matched sentence and extracted dependency triples for the aspects, yielding a total 3,252 cases across 782 Shorts videos (see Table~\ref{tab:dep-parse-counts}).

Dependency triples consist of the aspect, its syntactic head and the dependency role in the parse. We normalized surface forms of aspects and syntactic heads wrt.\ Unicode, punctuation, possessives, and plurals, using an entity lexicon.  Then, we attached sentence-level metadata (video id, seg group id, seg ids, start, end, sent ix). 

Consider the example in Figure~\ref{parser}, which illustrates the relevant dependency structure. The matched aspects, \textit{Netanyahu} and \textit{Israeli}, are shown in beige, their heads, \textit{wants} and \textit{regime}, in blue; the dependency relations \texttt{nsubj} and \texttt{amod} link aspect and head.

\subsection{Aspect-based Sentiment Analysis}
Since our domain for ABSA is different from existing datasets, we need to finetune an ABSA model on the targets of interest. Following \citet{miehling-dakota-kbler:2025:RANLP}, we apply a similar ABSA pipeline to our transcripts, using their gold standard as our base training set and extending it with additional validated instances for the newly introduced aspect targets. In contrast to user-generated comments, these transcripts are more syntactically well-formed and formal, which allows us to expand the range of aspect categories included in the analysis.

To annotate our new training examples, we use an end-to-end ABSA model (\texttt{DeBERTa-v3-large-absa-v1.1} \cite{YangPyABSA}) for all transcripts, retain 1{,}255 predictions with confidence $p \geq 0.75$, manually validate these cases, and merge the verified examples into the existing gold standard. Several aspect groups exhibit substantial class imbalance, most notably \textit{Arabs} and \textit{politicians}. The latter includes \textit{Israeli politicians} such as \textit{Netanyahu}, \textit{Smotrich}, \textit{Gallant}, and \textit{Ben-Gvir}, as well as opposing figures like \textit{Nasrallah}, \textit{Sinwar}, and \textit{Haniyeh} (see Table~\ref{tab:canonical-absa} for the resulting label distribution). However, we refrained from augmenting the dataset with synthetic examples to avoid skewing the empirical representation of politically charged aspect terms.

Using the augmented dataset, we finetuned several models under identical splits: \texttt{RoBERTa-base} \cite{liu2019roberta}, \texttt{DeBERTa-v3-base}, \texttt{DeBERTa-v3-large} \cite{he2021debertav3}, \texttt{DeBERTa-v3-large-absa-v1.1} \cite{YangPyABSA}, and a \texttt{Qwen2.5-7B-Instruct} LLM \cite{qwen2025qwen25technicalreport} finetuned with QLoRA \cite{dettmers2023qlora}.

\texttt{DeBERTa-v3-base} achieves the best performance on this structured, transcript-based political ABSA task (macro-F1 = 81.9), outperforming larger encoder variants, the ABSA-specialized Yang model, and the Qwen LLM (macro-F1 = 72.5) (see Appendix~\ref{appendix:configs} for all results).

\subsection{Video Frame Sampling}
To sample frames from the videos, we utilized VideoCapture from the \texttt{cv2} library. We obtained the frames per second per video and iterated over the frames of the video, saving one frame per second, 
resulting in a collection of one image per second of each video. 
Our sampling uses a uniform-FPS strategy, which aligns with \citet{brkic2025framesamplingstrategiesmatter}. The processing of all 2,371 Shorts was conducted at a fixed rate of one FPS across YouTube Shorts formats such as \texttt{.mp4}, \texttt{.webm}, \texttt{.mkv}, \texttt{.mov}, \texttt{.m4v}, and \texttt{.avi}. Each record includes the video title, an internal \texttt{video\_id}, and a \texttt{frame\_id} indicating the frame’s position within the original video. Overall, the dataset contains 2,861 frames from BBC, 3,957 from DW, 42,972 from Al Jazeera, and 44,252 from TRT.

\subsection{Semantic Scene Analysis (VLM-based)}

For visual classification, we employ the open-source \texttt{Qwen3-VL} model (4B) \cite{qwen3vl_2025}, which is the best option for image-text reasoning tasks given our limited computational resources.
The model assigns each frame to one of the seven semantic scene categories that capture recurrent visual patterns in conflict-related news video (see Appendix~\ref{appendix:prompt}). We refrained from using commercial systems such as Gemini~2.5~Pro and GPT-5, which may achieve stronger results on multimodal benchmarks \cite{comparebench_2025} but  do not provide reproducible results.

\section{Assessing Text \& Vision Performance}
To assess whether our pipeline supports robust multimodal analysis of short-form content, we conduct a manual inspection of language filtering, transcription quality, and ABSA on Shorts transcripts, followed by a quantitative evaluation of semantic scene classification using a held-out dataset.

\subsection{Transcription Quality and Language Filtering}
Although \texttt{Whisper} detects multiple languages, transcription quality deteriorates under noisy conditions such as chanting or overlapping voices, occasionally leading to errors in named entities and language identification. After an initial language check, we manually inspected randomly selected examples.

We observed such errors particularly in short, shouted utterances from rallies. In one video from a rally against Israel\footnote{\url{https://www.youtube.com/shorts/cS9KzCsYGds}}, the model misidentified the language and decoded the chant in Turkish (see~\ref{ex:katil}) instead of the intended text shown in~\ref{ex:katil1}.

\ex. \label{ex:katil} Geçin! Katil abd! Katil ismail! Katil abd! Katil ismail! Geçin! Geçin! Geçin!

\ex. \label{ex:katil1} Geçin! Katil ABD! Katil İsrail! Katil ABD! Katil İsrail! Geçin! Geçin! Geçin!

Here, \textit{İsrail} is mis-transcribed as \textit{İsmail}, reflecting the model’s reliance on acoustic similarity in short, high-intensity speech segments with limited linguistic context.

\subsection{Applying ABSA on Shorts Transcripts}
Our manual inspection indicates that \texttt{DeBERTa-v3-base} performs reliably on political communication tasks, including aspects with limited finetuning data. Qualitative evaluation of newly introduced aspects shows strong true positive rates for politically charged statements (see~\ref{ex:idf} and~\ref{ex:arab}).

\ex. \label{ex:idf}
The IDF will continue to operate according to international law.

\ex. \label{ex:arab}
Iran's axis of terror confronts America, Israel, and our Arab friends.

Both examples are correctly classified as \textit{positive}, demonstrating fine-grained distinctions on new aspect targets \textit{IDF} and \textit{Arabs}, despite their low-frequency representation in the training data.

\ex. \label{ex:hamas}
Hamas is not a terrorist group for us, of course, as you know.

\ex. \label{ex:denial}
There was no tradition of anti-Semitism in the Arab world.

Examples~\ref{ex:hamas} and \ref{ex:denial} are classified
as \textit{neutral}. \ref{ex:hamas} is ambiguous while 
\ref{ex:denial} reflects the model’s reliance on surface-level cues rather than implicit rhetorical functions. Such edge cases highlight known challenges for context-sensitive sentiment interpretation in political discourse.

However, we find that the model’s stable performance on structured, \texttt{Whisper}-generated transcripts supports using these predictions for longitudinal analysis across outlets and aspect categories.

\begin{table}[t]
\centering
\small
\begin{tabular}{lrrrr}
Category & True & False & Total & F. \% \\
\hline
combat / military       & 65 & 4  & 69  & 5.8 \\
destruct. / crisis & 69 & 12 & 81 & 14.8 \\
news / interview           & 89 & 2  & 91 & 2.3 \\
other / unknown                   & 165 & 63 & 228 & 27.6 \\
polit. / dipl.\_event     & 119 & 13 & 132 & 9.8 \\
public\_protest   & 54 & 2  & 56 & 3.6 \\
symb. / religious      & 133 & 9  & 142 & 6.3 \\
\hline
Total                      & 694 & 105 & 799 & 13.1 \\
\hline
\end{tabular}
\caption{Manual evaluation of semantic scene-type predictions for a held-out sample. (F. \%: \% of false frames)}
\label{tab:scene-eval}
\end{table}

\subsection{Performance on Semantic Scene Classification}
We manually evaluated 799 randomly sampled frame-level labels across all outlets in a held-out set to assess the model’s predictions. Table~\ref{tab:scene-eval} reports correct and incorrect classifications by scene type. \texttt{Qwen3-VL} achieved an overall accuracy of 86.9\% (694/799 frames).

\begin{figure}[t]
    \centering
    \includegraphics[width=.46\textwidth]{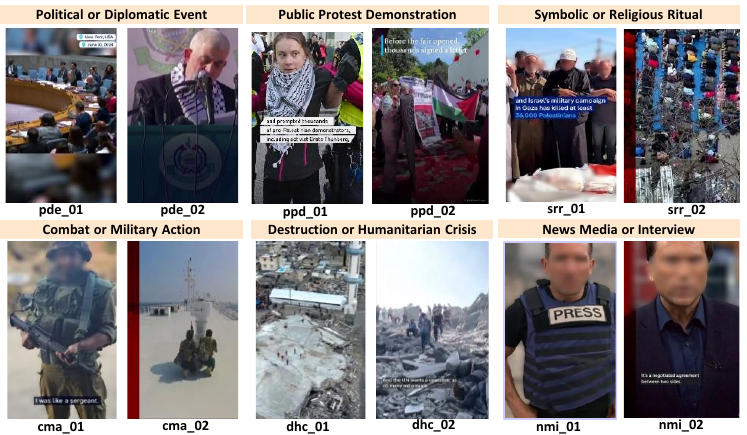}
    \caption{True-positive examples from scene classification (faces of non-publicly known individuals blurred).}
    \label{fig:vis}
\end{figure}

\begin{figure}[t]
    \centering
    \includegraphics[width=.46\textwidth]{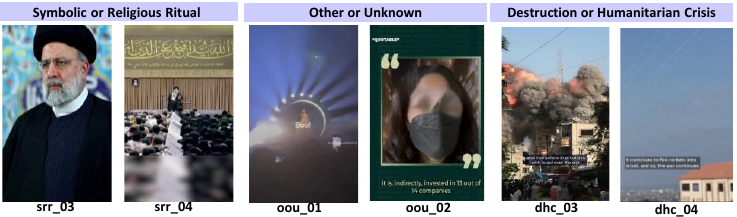}
    \caption{Examples of challenging cases in semantic scene classification.
    }
    \label{fig:edge}
\end{figure}

The model performs particularly well on \textit{news media / interview} scenes, with an accuracy of approximately 97\%. It reliably distinguishes reporter stand-ups (e.g., Fig.~\ref{fig:vis}, image nmi\_01) from visually similar \textit{destruction / humanitarian crisis} imagery (e.g., image dhc\_2), even in rapidly alternating sequences. The model also classifies \textit{political / diplomatic event} scenes consistently, which typically exhibit stable visual structures such as speeches, parliamentary settings, or UN sessions (Fig.~\ref{fig:vis}, images pde\_01 and pde\_02). Pop culture events, such as the Eurovision Song Contest, are correctly assigned to \textit{other / unknown} (Fig.~\ref{fig:edge}, image oou\_01), despite the presence of text overlays.

Errors are distributed unevenly across categories, with the highest false-positive rate occurring in \textit{other~/ unknown}, reflecting its broad semantic scope and visual ambiguity. A recurring error involves Al Jazeera’s ``Quotables,'' which are sometimes misclassified despite primarily representing \textit{news media / interview} content (Fig.~\ref{fig:edge}, image oou\_02). Figure~\ref{fig:edge} further illustrates ambiguous edge cases in which visual cues plausibly belong to multiple categories. For example, frames featuring Ebrahim Raisi are occasionally classified as \textit{symbolic / religious ritual} (images srr\_03 and srr\_04), although \textit{political / diplomatic event} would also be plausible. The model justifies these predictions with explanations such as ``religious figure in traditional attire,'' referencing Raisi’s \textit{qaba}. More broadly, we observe systematic overlap between \textit{combat / military action} and \textit{destruction / humanitarian crisis}, as damaged environments may indicate either active conflict or its aftermath.

Although the taxonomy necessarily simplifies the visual diversity of short-form news, our evaluation indicates that it reliably captures dominant semantic scene types. Most predictions fall into a small number of categories: \textit{news media / interviews} (34.7\%), \textit{destruction / humanitarian crises} (16.4\%), and \textit{political / diplomatic events} (11.8\%). Overall, \texttt{Qwen3-VL} and the proposed taxonomy provide a robust basis for comparative analysis of war-related short-form video content.

\section{Findings}

To address our \textbf{RQ}, we examine how sentiment and visual scene types vary across outlets and over time. We compare outlet-level sentiment distributions and monthly trends, relate sentiment cues to audience exposure using upload dates and view counts, and analyze differences in scene-type frequencies across outlets over the one-year period, including all 2{,}355 Shorts.

\begin{table*}[t]
\centering
\setlength{\tabcolsep}{2.5pt}
\begin{tabular}{cc}
\begin{minipage}{0.48\textwidth}
\centering
\textbf{Al Jazeera (AJ)}\\[1mm]
\begin{tabular}{lrrrr}
Aspect & Neg & Neut & Pos & Total \\
\hline
Arab & 0 & 23 & 2 & 25 \\
Gaza & 17 & 251 & 41 & 309 \\
Islam & 0 & 8 & 0 & 8 \\
Islamism & 19 & 112 & 3 & 134 \\
Israel & 302 & 371 & 19 & 692 \\
Israeli Politicians & 27 & 66 & 3 & 96 \\
Jews & 8 & 34 & 1 & 43 \\
Opp.\ non-Israeli polit. & 2 & 9 & 0 & 11 \\
Palestine & 7 & 184 & 97 & 288 \\
Zion & 4 & 4 & 0 & 8 \\
\hline
Overall & 386 & 1062 & 166 & 1614 \\
\hline
\end{tabular}

\vspace{2mm}

\textbf{BBC}\\[1mm]
\begin{tabular}{lrrrr}
Aspect & Neg & Neut & Pos & Total \\
\hline
Arab & 0 & 1 & 0 & 1 \\
Gaza & 2 & 51 & 0 & 53 \\
Islam & 0 & 0 & 0 & 0 \\
Islamism & 10 & 36 & 0 & 46 \\
Israel & 32 & 80 & 4 & 116 \\
Israeli Politicians & 0 & 1 & 0 & 1 \\
Jews & 0 & 1 & 0 & 1 \\
Opp.\ non-Israeli polit. & 0 & 0 & 0 & 0 \\
Palestine & 0 & 17 & 1 & 18 \\
Zion & 0 & 0 & 0 & 0 \\
\hline
Overall & 44 & 187 & 5 & 236 \\
\hline
\end{tabular}
\end{minipage}
&
\begin{minipage}{0.48\textwidth}
\centering
\textbf{Deutsche Welle (DW)}\\[1mm]
\begin{tabular}{lrrrr}
Aspect & Neg & Neut & Pos & Total \\
\hline
Arab & 0 & 2 & 0 & 2 \\
Gaza & 0 & 14 & 2 & 16 \\
Islam & 0 & 0 & 0 & 0 \\
Islamism & 3 & 23 & 0 & 26 \\
Israel & 13 & 50 & 0 & 63 \\
Israeli Politicians & 0 & 3 & 0 & 3 \\
Jews & 0 & 0 & 0 & 0 \\
Opp.\ non-Israeli polit. & 1 & 4 & 0 & 5 \\
Palestine & 0 & 9 & 4 & 13 \\
Zion & 0 & 0 & 0 & 0 \\
\hline
Overall & 17 & 105 & 6 & 128 \\
\hline
\end{tabular}

\vspace{2mm}

\textbf{TRT World}\\[1mm]
\begin{tabular}{lrrrr}
Aspect & Neg & Neut & Pos & Total \\
\hline
Arab & 0 & 10 & 3 & 13 \\
Gaza & 12 & 136 & 44 & 192 \\
Islam & 2 & 18 & 8 & 28 \\
Islamism & 11 & 53 & 2 & 66 \\
Israel & 205 & 154 & 48 & 407 \\
Israeli Politicians & 15 & 19 & 2 & 36 \\
Jews & 4 & 56 & 4 & 64 \\
Opp.\ non-Israeli polit. & 0 & 8 & 0 & 8 \\
Palestine & 20 & 205 & 200 & 425 \\
Zion & 18 & 12 & 5 & 35 \\
\hline
Overall & 287 & 671 & 316 & 1274 \\
\hline
\end{tabular}
\end{minipage}
\end{tabular}
\caption{Predicted sentiment distributions by aspect for all outlets.}
\label{tab:aspect_sentiment_two_column}
\end{table*}

\begin{figure}[t]
    \centering
    \includegraphics[width=.477\textwidth]{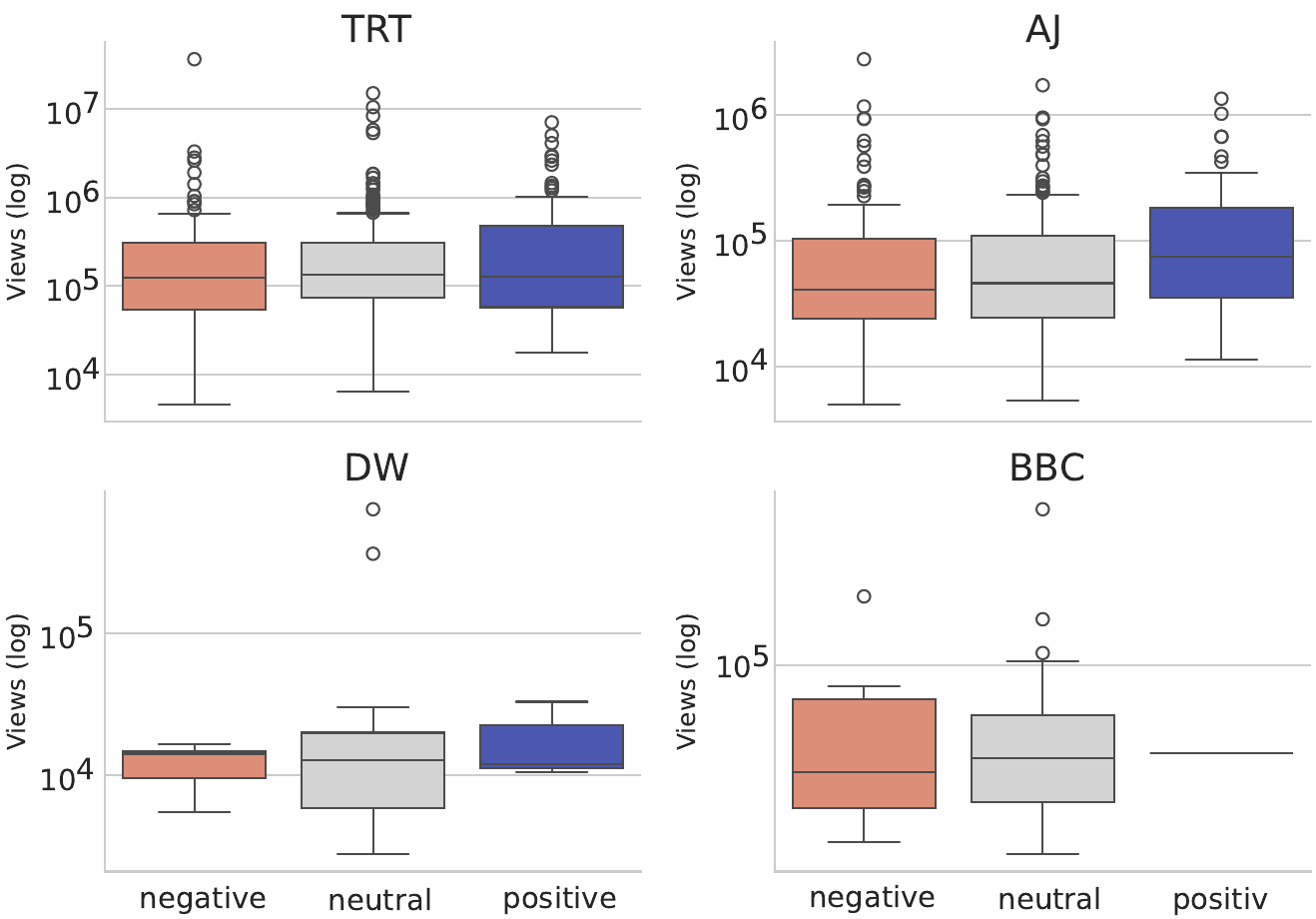}
    \caption{Sentiment comparison per outlet (log-transformed view counts).}
    \label{fig:box}
\end{figure}

\paragraph{Cross-Outlet Sentiment Patterns.}
Aggregated predictions reveal distinct outlet-specific affective toning (see Table~\ref{tab:aspect_sentiment_two_column}). Al Jazeera (AJ) and TRT display the strongest polarity, characterized by sustained negative sentiment toward \textit{Israel} and positive sentiment toward \textit{Palestine}-related aspects. In contrast, BBC and DW maintain a more neutral stance, though predictions for the latter should be interpreted cautiously given their smaller corpora. 
Across the 782 videos containing at least one aspect, 440 are neutral, 136 are positive, and 206 are negative, indicating that 43.7\% of transcripts exhibit a non-neutral emotional tone. Negative framings occur more frequently than positive ones, suggesting a general tendency toward negative affective cues across outlets.

We then examined these patterns in relation to view counts (Figure~\ref{fig:box}\footnote{Due to the highly skewed nature of engagement distributions, we analyzed log-transformed view counts to reduce the influence of a small number of extremely viral videos.}). Positive sentiment yields the highest median views for AJ and TRT; BBC shows the opposite pattern, with negatively framed videos receiving higher view counts. DW exhibits lower and more homogeneous view counts overall.

\begin{figure}[t]
    \centering
    \includegraphics[width=.46\textwidth]{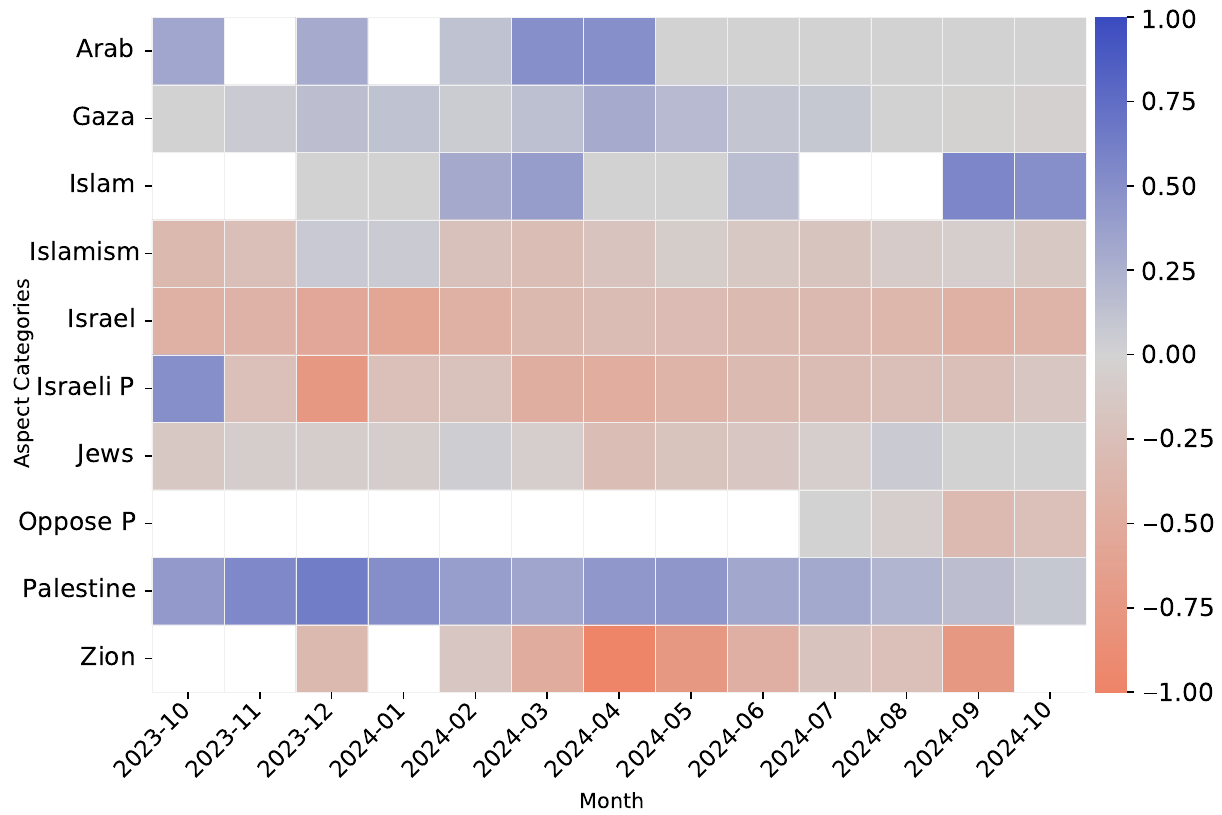}
    \caption{Sentiment trends by aspect category over time. Colors indicate polarity: pos. (blue), neut. (grey), neg. (red)}
    \label{fig:heat}
\end{figure}

\paragraph{Sentiment Dynamics Over Time.}
Next, we looked at temporal trends in Figure~\ref{fig:heat}, which shows each month during the collected period separately. This analysis shows persistent negative framing of \textit{Israel} and consistently positive framing of \textit{Palestine}, a pattern stable across the full year of data. For the latter, positive sentiment surged in transcripts during December 2023, including videos of activists storming the U.S.\ Senate to protest weapon transfers to Israel. In these cases, the transcripts capture chants such as \textit{Free Palestine} from protesters detained inside the building while spoken narration from the outlet itself is absent. This trend is also reflected in visual frames from the corresponding videos, which are classified as \textit{public protest~/ demonstration}.

Mentions of \textit{Zionism} remain consistently negative over time while e.g., \textit{Islam}-related terms stay largely neutral in most instances. We observe that transcripts with explicit emotional tone often reproduce on-scene signals, such as protest chants or quotes from interviewees voicing partisan views, through these videos. Example~\ref{ex:propaganda} illustrates this pattern and reflects the strongly negative framing associated with \textit{Zionist}.

\ex. \label{ex:propaganda}The majority of Israelis support this Zionist, genocidal, apartheid, racist, colonial takeover of all Palestine, with extreme brutality to civilians.

Such sentences rarely originate from news anchors within transcripts, yet their persuasive power is reinforced by consistent embedding in highly partisan news coverage.

Barely any cases exist in which Israeli politicians are positively evaluated; only five instances in total, shortly after October 7. The few positive cases typically originate from interviews with other political leaders. For example,~\ref{ex:netanyahu} quotes former U.S. President Biden speaking about PM Netanyahu in the context of the war in Gaza.

\ex. \label{ex:netanyahu}In a constructive meeting with Prime Minister Netanyahu today, he confirmed that Israel accepts the bridging proposal and supports it.

\begin{table}[t]
\centering
\small
\begin{tabular}{lrrrr}
Category & AJ & BBC & DW & TRT \\
\hline
combat / military            & 2071 & 208  & 295 &  1920 \\
destruct./ crisis   & 6116 & 445  & 580 & 8255   \\
polit. / dipl.\_ev.      & 5315 & 137  & 713  & 7585  \\
news/ interview    & 17468 & 1223  &  906 & 13070  \\
public\_protest      & 4345 & 184  &  531 & 6702  \\
symb. /religious         & 1382 & 48  & 114  & 1901 \\
other / unknown                      & 6275 & 616  & 713 & 7585 \\
\hline
Video IDs & 924 & 68 & 85 & 1258 \\
\hline
\end{tabular}
\caption{Distribution of semantic scene types.}
\label{tab:umbrella-outlet}
\end{table}

\begin{figure}[t]
    \centering
    \includegraphics[width=.47\textwidth]{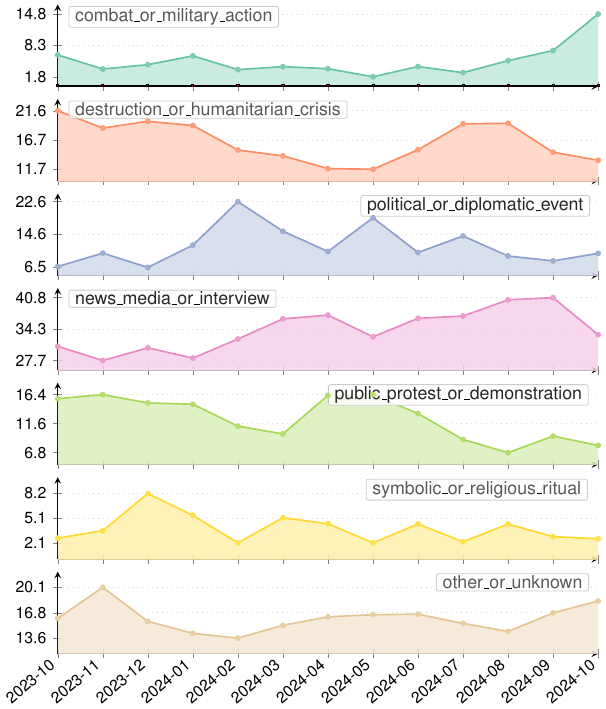}
    \caption{Semantic scene types over time (monthly percentages by category adjusted y‑axis per row).}
    \label{fig:area}
\end{figure}

\paragraph{Cross-Outlet Visual Patterns.}
We applied our new taxonomy to video frames sampled at one frame per second. Table~\ref{tab:umbrella-outlet} shows the distribution of semantic scene types across outlets. Because all videos were manually selected, the corpus reflects digital discourse surrounding the Israel–Hamas war during 2023–2024. To provide context, we aligned frame-level spikes with real-world timelines and examined corresponding events.

Figure~\ref{fig:area} shows the semantic scene types over time (sampled per month). \textit{News media / interview} accounts for the largest share at 34.7\%, reflecting anchor-led segments and correspondent interviews, with a noticeable increase toward the end of the period. The second largest category, \textit{destruction / humanitarian crisis} (16.4\%), peaks in the early stages of the war and resurfaces between July and September 2024, coinciding with the Khan Yunis crisis and a polio outbreak. The July peak corresponds to footage of an orphaned Palestinian child, which received 6.1 million views, the second-highest view count among Shorts in July and August. The August peak is associated with footage of destroyed buildings and civilians carrying wounded individuals, generating over 2.9 million views, the third-highest view count during this period. Notably, these Shorts include text overlays but lack spoken narration from news outlets.

\textit{Other / unknown} represents 16.2\%, with a spike in November 2023 linked to hostage deals, prisoner swaps, and later cultural events such as Eurovision-related controversies. Frames classified as \textit{public protest / demonstration} make up 12.5\%, emerging shortly after October 7 and intensifying during campus encampments in April and May, including coverage of protests at Columbia University (34 Shorts), Emory University, and a major demonstration in Paris. The footage consistently highlights scenes of violence, depicting not only police crackdowns, but also protester violence.

\textit{Political / diplomatic events} account for 11.8\% of all frames. Peaking in February, this category corresponded to footage from UN assemblies and other high-level diplomatic meetings (21 shorts), debates over the Rafah ground offensive (18 shorts), and International Court of Justice (ICJ) hearings addressing Israel’s occupation, as well as accusations of apartheid and genocide (eight shorts). \textit{Combat~/ military action} is comparatively rare at 4.8\%, yet its share rises sharply toward the end of the timeline, indicating that combat footage becomes more frequent in October 2024. The smallest category, \textit{symbolic / religious ritual}, remains at 3.7\%, appearing mainly in frames of vigils and Friday prayers. 

Overall, our findings demonstrate that visual discourse mirrors real-world developments and accurately predicts semantic scene types over time.

\subsection{Discussion}
Applying our pipeline to the dataset yields mixed insights. On the textual side, ABSA captures sentiment polarity in short-form transcripts, revealing that coverage is often strongly polarized. State-funded outlets frequently embed emotionally charged framings in quotes from interviewees or on-site signals such as protest chants, with particularly persistent negative sentiment toward \textit{Israel} and positive sentiment toward \textit{Palestine} over time. Videos with stronger polarity—especially AJ and TRT’s positive framing of Palestinians—tend to achieve higher median view counts, while BBC’s negatively framed segments attract more views. These findings align with prior work documenting partisan imbalance in digital discourse of the war \cite{alamsyah2024, miehling-dakota-kbler:2025:RANLP, steffen2025memesmultimodaltopicmodeling}, suggesting that short-form formats may amplify, rather than mitigate, polarization through selective emotional cues.

On the visual side, we processed over 94{,}000 frames using a small semantic taxonomy, which classified approximately 84\% of the material, leaving 16\% in the residual \textit{other / unknown} category. Despite its small number of categories, the taxonomy captures meaningful distinctions in conflict coverage. For example, TRT includes \textit{destruction / humanitarian crisis} imagery in 19\% of its videos, followed by BBC at 16\%, while 15\% of TRT’s frames depict \textit{public protest / demonstration} scenes that carry high emotional weight. Tracing these visual patterns over time provides important context for interpreting sentiment dynamics and audience engagement, although automated classification alone cannot replace qualitative interpretation.

While prior work suggests that conflict reporting often aligns textual framing with visual cues to influence audience perception \cite{chatterje-doody2019making}, our analysis of short-form news videos indicates that such alignment is rarely overt. Instead, affective cues emerge through a subtle interplay of audio, narrative structure, text overlays, and visual framing, most clearly in protest scenes featuring chanting or explicit expressions of distress. Overall, affective cues are interwoven in subtle ways, making it difficult to identify clear bimodal patterns that may increase the potential of audience polarization--even in partisan-driven news content.

\section{Conclusion \& Future Work} 
We introduce a multimodal pipeline based on open-source models for analyzing political communication on YouTube Shorts. By combining transcript-based ABSA with VLM-based semantic scene classification, we generate robust insights through the joint analysis of textual and visual cues.

Our evaluation further shows that smaller models can outperform large transformers and LLMs when domain adaptation is prioritized.
This finding underscores the value of task-specific finetuning over  model size for achieving highly accurate results in domain-specific settings.

The proposed pipeline offers a practical solution for researchers interested in monitoring digital discourse across platforms such as YouTube, TikTok, and Instagram. While current state-of-the-art models often demand technical resources beyond the reach of most humanities scholars, our method provides a resource-efficient alternative without sacrificing analytical depth. Ultimately, this approach enables a multifaceted analysis of highly dynamic online environments, supporting future research on polarization, rhetorical framing, and visual cues in social media discourse.

Future work will focus on tighter integration of ABSA and semantic scene analysis at the outlet level, which requires a larger per-outlet video base. We are also planning to expand this work to other platforms, such as TikTok, and other languages.

\section{Limitations}
While the content was manually pre-selected, the pipeline demonstrates an effective approach for consistently distinguishing and classifying semantic scene types across outlets using a simplified taxonomy for war-related coverage and surrounding discourse. We observed some irregularities, despite the model’s ability to make fine-grained distinctions between scene types. 
The category \textit{other~/ unknown} led to the most false positives. This is likely due to its brief prompt description, which leaves too many loopholes for misclassification. A more elaborate and precise prompt has the potential to yield more accurate results by reducing ambiguity and narrowing the semantic scope of this residual category.

This study focuses on English-language content. However, all four news outlets also publish YouTube content in other languages. Therefore, results based on English-language corpora may differ from analyses of non-English content from the same period. Thus, our results may not fully reflect the overarching narratives of these outlets on YouTube.

Although frames were classified robustly, the presence of AI-generated imagery within videos raises concerns about whether such content reflects authentic discourse or fabricated material. 

\section{Ethics Statement}
Our study was conducted in accordance with institutional IRB approval and ethical research standards. We show content that includes sensitive and potentially harmful language and visual frames, such content may endorse or glorify violence. These examples are presented solely for illustrative purposes and are necessary to address our research questions. We acknowledge that misuse of the proposed pipeline could facilitate the dissemination of biased or polarizing content, which may negatively impact individuals and minority groups targeted by such statements.

\section{Acknowledgments}
This work used Jetstream2\footnote{https://jetstream-cloud.org/} at Indiana University through allocation HUM200003 from the Advanced Cyberinfrastructure Coordination Ecosystem: Services \& Support (ACCESS)\footnote{https://access-ci.org/} program, which is supported by National Science Foundation grants \#2138259, \#2138286, \#2138307, \#2137603, and \#2138296. 
We also thank Daniel Dakota for his help with parsing and Stephen Zhong for his help in running initial VLM experiments.

\section{Bibliographical References}\label{sec:reference}
\bibliographystyle{lrec2026-natbib}
\bibliography{lrec2026-example}

\clearpage

\appendix

\onecolumn

\section{Appendix}

\subsection{Annotation Prompts}
\label{appendix:qlora}
\paragraph{Qwen2.5-7B}
\begin{small}
\begin{verbatim}
You are a domain expert in political communication performing aspect-based sentiment analysis. 
Sentence: \{TEXT\} Aspect: \{ASPECT\} 
Answer with exactly one word: negative, neutral, or positive.
\end{verbatim}
\end{small}

\paragraph{Qwen3-VL (4B)}
\label{appendix:prompt}

During the experimental phase, we iteratively refined the semantic scene taxonomy by prompting \texttt{Qwen3-VL} with an initial set of 14 scene types, followed by 12 and then 7 types, across approximately 4{,}000 sampled frames. For each frame, the model produced a structured JSON object containing the predicted scene type, a binary text overlay and short evidence phrases describing visible cues. 
Evidence suggested that the model struggled to distinguish between the initial categories (e.g. civilian\_life\_under\_crisis vs. refugee\_or\_displacement) due to shared visual cues. Therefore, we merged such categories into broader scene types (e.g. destruction or humanitarian crisis) and expanded the description of the prompt.

\begin{small}
\begin{verbatim}
Semantic scene types and their descriptions based on visually observable cues

You are a careful visual annotator. Look only at the image.

Decide what kind of scene is shown using the categories below. Be conservative:
if unclear, set "abstain": true or use "scene_type": "other_or_unknown". 
Output JSON ONLY.

Scene type options (decide based on visible cues):

- "combat_or_military_action": Weapons, explosions, airstrikes, armed soldiers 
in action or at checkpoints.

- "destruction_or_humanitarian_crisis": Rubble, collapsed buildings, smoke, 
  damaged streets, tents or shelters, refugees, queues for aid, doctors or
  rescuers helping civilians.

- "political_or_diplomatic_events": Politicians or officials at podiums 
  or in formal meetings, parliaments, press rooms, negotiation tables, 
  government ceremonies.

- "news_media_or_interview_settings": TV anchors in a studio, reporters 
  speaking to camera, people being interviewed  with a microphone, 
  talk show or split screen news formats.

- "public_protest_or_demonstration": Crowds holding signs or banners, 
  marches, rallies, vigils in streets or squares, police lines facing 
  demonstrators (including protests outside the conflict region).

- "symbolic_or_religious_ritual": Religious buildings or interiors,
  prayer, clergy, funerals, coffins, memorials, monuments, candlelight
  vigils, large flags used ceremonially or symbolically.

- "other_or_unknown": Any scene that does not clearly 
  match the above categories or is too ambiguous.
\end{verbatim}
\end{small}

\pagebreak

\subsection{Lexicon for ABSA-linking}\label{appendix:lexicon}

\begin{table*}[h]
\centering
\small
\setlength{\tabcolsep}{4pt}
\renewcommand{\arraystretch}{1.05}
\begin{tabular}{lllllllll}
Jews & Zion & Israel & Islamism & Islam & Palestine & Israeli\_P & Oppose\_P & Arab \\
\hline
Jews     & Zionists   & Israel     & Hamas       & Muslims     & Palestine      & Netanyahu   & Haniyeh    & Arab \\
Jewish   & Zionist    & Israeli    & Hezbollah   & Muslims'    & Palestinians   & Gvir        & Sinwar     & Arabs \\
Jew      & Zionist’s  & Israelis   & Hamas's     & Muslim      & Palestine's    & Smotrich    & Abbas      & Arabic \\
Jew’s    & Zionists’s & Israel's   & Hezbollah’s & Muslims’s   & Palestinian    & Gallant     & Nasrallah  & Arabics \\
Judaism  & Zionism    & IDF        & Jihad       & Islam       & Palestinian’s  & Netanyahu's & Nasrallah’s & Arabic’s \\
         &            &            & jihadists   &             &                & Gallant's   & Abbas’s    & \\
         &            &            & jihadist    &             &                & Smotrich's  & Sinwar’s   & \\
         &            &            & martyr      &             &                & Ben-Gvir’s  & Haniyeh’s  & \\
         &            &            & martyrs     &             &                &             &            & \\
         &            &            & shahid      &             &                &             &            & \\
\hline
\end{tabular}
\caption{Aspect Lexicon Category for ABSA.}
\label{tab:lexicon-absa}
\end{table*}

\subsection{ABSA Fine-tuning configuration}
\label{appendix:configs}
\begin{table}[h]
\centering
\small
\begin{tabular}{lrrrr}
Model & LR & Ep. & Batch & F1  \\
\hline
RoBERTa-base & 3e$^{-5}$ & 6 & 16 & 76.32  \\
DeBERTa-v3-base & 3e$^{-5}$ & 6 & 16 & \textbf{81.89}  \\
DeBERTa-v3-large & 1e$^{-5}$ & 4 & 8 & 78.43  \\
DeBERTa-v3-large-absa-v1.1 & 5e$^{-6}$ & 5 & 8 & 79.65\\
Qwen2.5-7B (QLoRA) & 2e$^{-5}$ & 3 & 4 & 72.50  \\
\hline
\end{tabular}
\caption{ABSA Fine-tuning configuration and evaluation based on best results for all models.}
\label{tab:absa-results}
\end{table}

\bibliographystylelanguageresource{lrec2026-natbib}
\bibliographylanguageresource{languageresource}

\end{document}